# Physical Deep Learning with Biologically Plausible Training Method

Mitsumasa Nakajima[1], Katsuma Inoue[2], Kenji Tanaka[1], Yasuo Kuniyoshi[2, 3], Toshikazu Hashimoto[1], and Kohei Nakajima[2, 3]

1. NTT Device Technology Labs., 3-1 Morinosato-Wakamiya, Atsugi, Kanagwa, 243-0198, Japan

2. Graduate School of Information Science and Technology, The University of Tokyo, 7-3-1 Hongo, Bunkyo-ku, Tokyo, 113-8656, Japan

3. Next Generation Artificial Intelligence Research Center, The University of Tokyo, 7-3-1 Hongo, Bunkyo-ku, Tokyo, 113-8656, Japan

**Abstract**

The ever-growing demand for further advances in artificial intelligence motivated research on unconventional computation based on analog physical devices. While such computation devices mimic brain-inspired analog information processing, learning procedures still relies on methods optimized for digital processing such as backpropagation. Here, we present physical deep learning by extending a biologically plausible training algorithm called direct feedback alignment. As the proposed method is based on random projection with arbitrary nonlinear activation, we can train a physical neural network without knowledge about the physical system. In addition, we can emulate and accelerate the computation for this training on a simple and scalable physical system. We demonstrate the proof-of-concept using a hierarchically connected optoelectronic recurrent neural network called deep reservoir computer. By constructing an FPGA-assisted optoelectronic benchtop, we confirmed the potential for accelerated computation with competitive performance on benchmarks. Our results provide practical solutions for the training and acceleration of neuromorphic computation.

**Main**

Machine learning based on artificial neural networks (ANNs) has successfully demonstrated its excellent ability through record-breaking performance on image processing, speech recognition, game playing, and so on [1]-[3]. Although these algorithms resemble the workings of the human brain, they are basically implemented on a software level using conventional von Neumann computing hardware. However, such digital-computing-based ANNs are facing issues regarding energy consumption and processing speed [4]. These issues have motivated the implementation of ANNs using alternative physical platforms[5], such as spintronic [6]-[8], ferroelectric [9], [10], soft-body [11,12], photonic hardware [13]-[17], and so on [18]-[21]. Even passive physical dynamics, interestingly, can be used as a computational resource in a randomly connected ANNs. This framework is called physical reservoir computing (RC) [20]-[22] or an extreme learning machine (ELM) [23], [24], whose ease of implementation has greatly expanded the choice of implementable materials and its application range. Such physically implemented neural networks (PNNs) enable the outsourcing of the computational load for specific tasks to a physical system such as a memory [25], optical link [26], [27], sensor component [28], [29], or robotic body [30]. The experimental demonstrations of these unconventional computations have revealed performance competitive with that of conventional electric computing [31]-[33].

Constructing deeper physical networks is one promising direction for further performance improvement because they extend network expression ability exponentially [34], [35], as opposed to the polynomial relationship in wide (large-node-count) networks. This has motivated proposals of deep PNNs using various physical platforms [14], [16], [28], [36]-[40]. Their training, so far, basically relies on a method called backpropagation (BP), which has seen great success in the software-based ANN. However, BP is not suitable for PNNs in the following respects. First, the physical implementations of the BP operation are still complex and unscalable [37]-[41]. Thus, the calculation for BP for a PNN is typically executed on an external regular computer with a simulation model of a physical system [14], [16], [28], [36], [40]. This strategy results in a loss of any advantage in speed or energy associated with using the physical circuit in the training process. Thus, this method is not suitable for *in-situ* (online) training; it is only usable for "train once and infer many times" application. Second, BP requires accurate knowledge about the whole physical system. Thus, the performance of the PNNs entirely relies on the model representation or measurement accuracy of the physical system [42]. In addition, when we apply BP to RC, these requirements spoil the unique features for physical RC, such as black-boxification of passive physical dynamics as a random network.

Similar to BP in the PNNs, the operational difficulty of BP in biological neural networks has also been pointed out in the brain science field; the plausibility of BP in the brain — the most successful analog physical computer — has been doubted [43]-[45]. These considerations have motivated the development of biologically plausible training algorithms [46]-[49]. One promising recent direction is direct feedback alignment (DFA) [50]. In this algorithm, fixed random linear transformations of the error signal at the final output layer are employed instead of the backward error signals. Thus, this approach does not require error BP nor knowledge of the weight. In addition, it was reported that the DFA scales to modern large-scale network models [51]. The success of such biologically motivated training suggests that there is a more suitable way to train PNNs than BP.

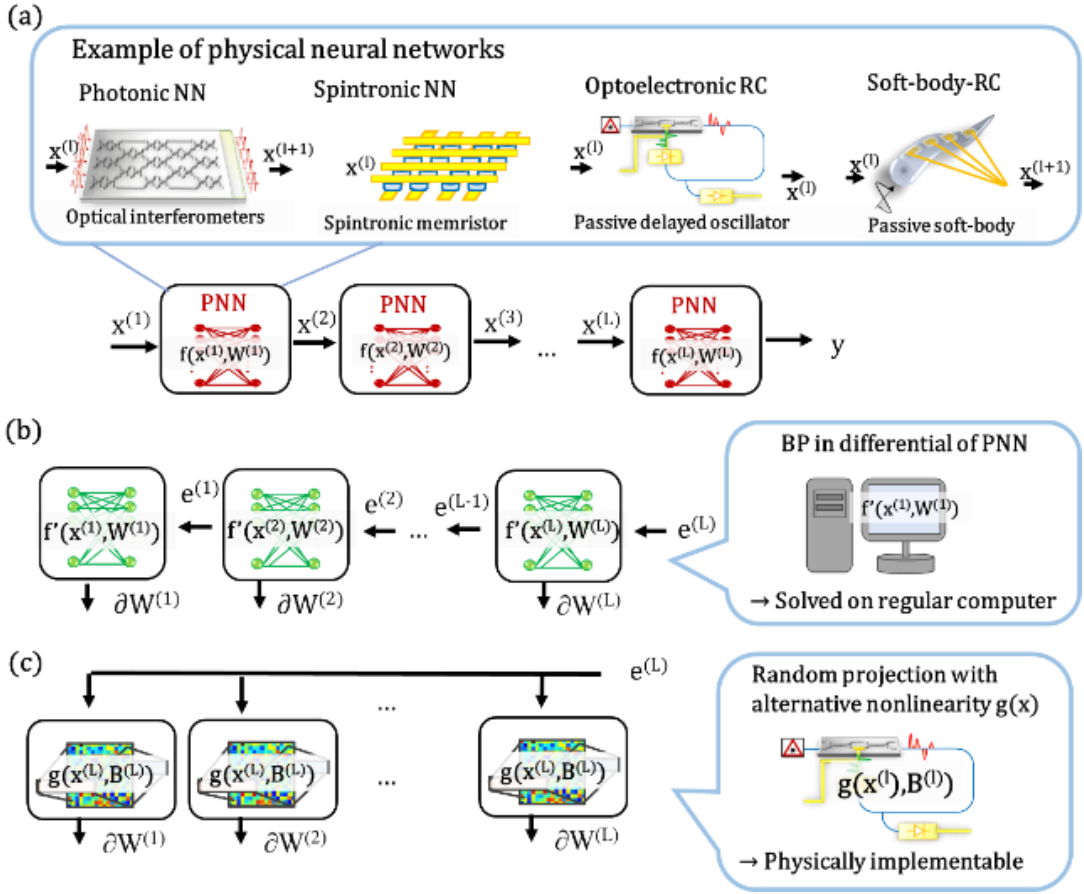

**Fig. 1.** *(a) Schematics of physical neural networks (PNNs). Training sequence of PNN with (b) backpropagation, and (c) augmented biologically plausible training called direct feedback alignment (DFA). Augmented DFA enables parallel, scalable, and physically accelerable training of deep physical networks based on random projection with alternative nonlinearity g(x).*

In this paper, we demonstrate physical deep learning by augmenting the DFA algorithm. While the standard DFA still requires the differential of physical nonlinear activation $f'(x)$, we replace it with arbitrary nonlinearity $g(x)$ and show that the performance is robust to the choice of $g(x)$. Thanks to this augmentation, we no longer need to simulate $f'(x)$ accurately. As the proposed method is based on parallel random projection with arbitral nonlinear activation, we can execute the computation for the training on a physical system in the same manner as with the physical ELM or RC concept [20]-[22]. This enables the physical acceleration of both inference and training. To demonstrate the proof-of-concept, we constructed a benchtop of FPGA-assisted optoelectronic deep physical RC. Although our benchtop is simple and easy to apply to various physical platforms with only software-level updates, we achieved performance comparable to that of large-scale complex state-of-the-art systems. Moreover, we compared the whole processing time, including that for digital processing, and found the possibility of physical acceleration of the training procedure. Our approach provides a practical alternative solution for the training and the acceleration of large-scale neuromorphic physical computation.

**Results**

**Physical deep learning with biologically plausible training** Figure 1(a) shows the basic concept of PNNs. The forward propagation of a standard multilayer network is described as $\mathbf{x}^{(l+1)} = f(\mathbf{W}^{(l)}\mathbf{x}^{(l)})$, where $\mathbf{W}^{(l)}$ and $\mathbf{x}^{(l)}$ are the weight and input tensor for the $l$th layer, and $f$ denotes nonlinear activation. In the PNN framework, this operation is executed on a physical system; i.e., $\mathbf{x}^{(l)}$, $\mathbf{W}^{(l)}$, and $f$ correspond to the physical inputs (e.g., optical intensity, electric voltage, vibration), physical interconnections (e.g., optical, electrical, or mechanical coupling) in the physical system, and physical nonlinearity (e.g., nonlinear optical/magnetic/mechanical effects), respectively. To train such networks, we need to update $\mathbf{W}$ to reduce given cost function $E$. A general solution is the BP algorithm shown in Fig. 1(b). The weight update rule for BP is obtained through the chain-rule as follows:

$$\delta \mathbf{W}^{(l)} = \frac{\partial E}{\partial \mathbf{W}^{(l)}} = -\mathbf{x}^{(l),T}[\mathbf{W}^{(l+1),T}\mathbf{e}^{(l+1)}] \odot f'(\mathbf{x}^{(l)}), \qquad (1),$$

where $\mathbf{e}^{(l)}$ is the error signal at the $l$th layer, defined as $\mathbf{e}^{(l)} = \partial E/\partial \mathbf{x}^{(l)}$, the superscript T denotes transportation, and $\odot$ denotes the Hadamard product. The training using eq. (1) is typically executed on an external regular computer by constructing a physical simulation model[14], [16], [28], [36], [40], which requires large computational cost. Thus, this strategy is not suitable for *in-situ* training. In addition, the error in the simulation model significantly affects PNN performance. Therefore, the training method for the PNN is still under considerations despite the success of BP in the software-based ANN.

Let us consider DFA as an alternative solution [see Fig. 1(c)]. In the standard DFA framework, the weighted gradient signals in eq. (1) are replaced with a linear random projection of the error signal at the final layer $L$ [50], [51]. Then, we can obtain the following update rule:

$$\delta \mathbf{W}^{(l)} = -\mathbf{x}^{(l),T}[\mathbf{B}^{(l),T}\mathbf{e}^{(L)}] \odot f'(\mathbf{x}^{(l)}), \qquad (2)$$

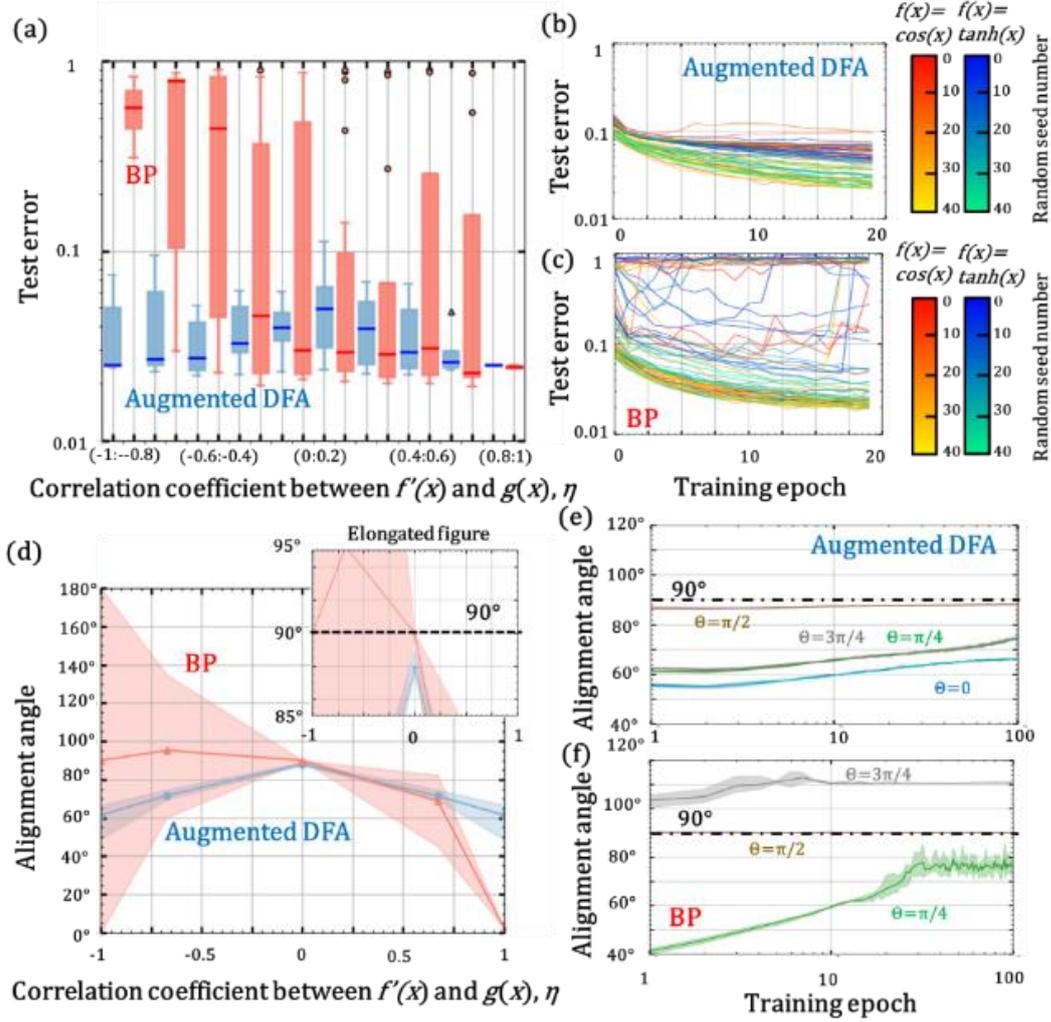

*Fig. 2. (a) Test error distribution of four-layer fully connected neural network as a function of the correlation coefficients between f'(x) and g(x), η. Blue and red boxplots in (a) are the results for the model trained by BP and augmented DFA. η was scanned by using various g(x):sin(x), cos(x), triangle(x), tanh(x) functions and random Fourier series. The boxes in the figure indicate the first and third quartile of the data distribution. The whiskers mean the minimum and maximum values except the outlier. Training curve for (b) augmented DFA and (c) BP with random Fourier series. Color differences indicate differences in activation function [f(x)=cos(x), tanh(x)] and differences in the random seed to generate g(x). (d) Averaged alignment angle under the training as a function of η. The colored region expresses the range within the minimum and maximum values. Evolution of alignment angle for layer 2 under training for (e) augmented DFA and (f) BP. Each experiment was repeated five times.*

where $\mathbf{B}^{(l)}$ is a random projection matrix for the *l*th layer update, and *f'* denotes the gradient of *f*. As shown in eq. (2), we can estimate the gradient without the information about $\mathbf{W}^{(l)}$. In addition, physical implementation of random projection process $\mathbf{B}^{(l),T}\mathbf{e}^{(L)}$ can be easily implemented by using various devices because this process is the same as the physical ELM and RC approach. By using commercially available photonic components, we can emulate $5 \times 10^5$ by $5 \times 10^5$ matrix operations on the single integrated optics [52], which are enough for the single hidden layer even in the state-of-the-art model. Note that the photonic acceleration of this process has already been demonstrated [53]. In addition, the DFA process can be parallelized because it is not a sequential equation unlike the one for the BP. Despite its simplicity, DFA can scale modern neural network models [51]. However, $f'(x)$ still remains in eq. (2), requiring accurate modeling and simulations, which is the bottleneck in the learning process for PNNs.

Here, we replace the $f'(x)$ function with the function $g(x)$ to investigate the robustness against the choice of $g(x)$. Then, we derive the update rule as

$$\delta \mathbf{W}^{(l)} = -\mathbf{x}^{(l),\mathrm{T}} \big[\mathbf{B}^{(l),\mathrm{T}} \mathbf{e}^{(L)}\big] \odot g(\mathbf{x}^{(l)}), \tag{3}$$

which is named *augmented DFA*. As the $f'(x)$ is replaced with $g(x)$, the equation no longer includes the knowledge for the parameters in the forward propagation. The gradient $\delta \mathbf{W}^{(l)}$ can be estimated from the final error $\mathbf{e}^{(l)}$ and alternative nonlinear projection of given input $\mathbf{x}^{(l)}$. As shown in the following section, we can select $g(x)$ almost arbitrarily. The only requirement is to avoid the function uncorrelated to $f'(x)$. Thus, the computation of $g(x)$ is also implementable to a physical system. Importantly, this treatment is also useful for black-box fixed physical networks such as physical RC, where black-box means that we do not know (or only have rough information about) $\mathbf{W}^{(l)}$ and $f$. When we apply the BP algorithm to physical RC, we need to simulate the gradient of the physical system using backpropagation through time (BPTT) using a regular computer. Thus, we need to open the black-box (measure and model $\mathbf{W}^{(l)}$ and $f$) to estimate the gradients. They spoil the advantage of such a randomly fixed physical network. On the other hand, augmented DFA can train the model without BPTT and knowledge of the physical system. Although we need to estimate an additional term compared with DFA for feedforward network [see eqs. (6)-(10) in Methods], this can be executed on physical hardware in the same manner as forward propagation in RC. Thus, we can improve the performance of RC while maintaining its unique feature. The detailed update rule is described in the method section and the concrete experimental demonstration is shown in the following section.

**Basic characterization of augmented DFA**    First, we investigated the effect of the augmentation of DFA, that is, the effect of $g(x)$, by using the standard image classification benchmark called Modified National Institute of Standards and Technology database (MNIST) task. In the experiment, the model was composed of four fully connected layers with 800 nodes for each layer and two types of nonlinear activation $f(x)$: hyperbolic tangent (tanh) and cosine (cos) function. These nonlinearities correspond to the simple model of commonly used photonic implementation. In this experiment, we generated $g(x)$ from some well-used functions (sine, cos, tanh, triangle) and the following random Fourier series: $g(x) = c_1 + \sum_{k=1}^{N} a_n \sin(kx) + b_n \cos(kx)$, where $a_k$, $b_k$, and $c_1$ are the random uniform coefficients sampled from $\mathbb{R} \in [-1:1]$. $N$ was set to 4 and normalized by the relationship $|c_1| + \sum_{n=1}^{N} |a_n| + |b_n| = 1$. Forty random Fourier series were examined in this experiment. To discuss the characterization quantitatively, we measured the correlation coefficient $\eta$ between $f'(x)$ and $g(x)$ defined in the method section.

Figure 2(a) shows the relationship between the correlation coefficient $\eta$ and the test error after the training. For comparison, we also show the results using the BP algorithm in eq. (1) with $f'(x)$ replaced with $g(x)$. The boxes in the figure indicate the first and third quartile of the data distribution. The whiskers mean the minimum and maximum values except the outlier. The outliers are also plotted in the figure as closed circles and triangles. The case for $\eta = 1$ means that $g(x)$ equals $f'(x)$, which corresponds to the case of the standard BP and DFA in eqs. (1) and (2). The cases for $\eta = 0$ and -1 correspond to uncorrelation and inverse correlation, respectively. The typical training curves for the augmented DFA and BP are shown in Fig. 2(b) and (c). The test error of BP increases sharply when $\eta$ deviates from 1 [see Fig. 2(a)], and it diverges in the various cases [see Fig. 2(c)]. In particular, almost no meaningful training was possible when the correlation coefficient was negative. This is because the update direction became opposite from the correct direction. On the other hand, the test accuracy for augmented DFA showed a gentle dependency on $\eta$ [see Fig. 2(a)], and it always converges [see Fig. 2(b)]. These results indicate that the training is highly robust to the choice of $g(x)$. The test error was maximized at $\eta = 0$ [i. e., $g(x)$ became the inverse function of $f'(x)$]. Even in such a case, we could obtain an accuracy of about 89%, far superior to that for the BP. In addition, learning can be performed when the correlation coefficient is negative. We think that this is due to the random weight matrix **B**; that is, the randomly distributed linear projection term in eq. (3) erases the positive/negative sign of $g(\mathbf{x})$.

One index for evaluating the feedback alignment algorithm well operated or not is the alignment angle ($\angle \boldsymbol{\delta}_{\text{BP/DFA}}$), which is the angle between $\boldsymbol{\delta}_{\text{BP}}$ and $\boldsymbol{\delta}_{\text{DFA}}$, where $\boldsymbol{\delta}_{\text{BP}}$, $\boldsymbol{\delta}_{\text{DFA}}$, and $\angle \boldsymbol{\delta}_{\text{BP/DFA}}$ are defined as follows: $\boldsymbol{\delta}_{\text{BP}} = \mathbf{W}^{\mathbf{T}} \mathbf{e}^{(l)}$ and $\boldsymbol{\delta}_{\text{DFA}} = \mathbf{B}^{(l)} \mathbf{e}^{(L)}$, $\angle \boldsymbol{\delta}_{\text{BP/DFA}} = \cos^{-1}(\boldsymbol{\delta}_{\text{BP}} \cdot \boldsymbol{\delta}_{\text{DFA}})$ [49]. When the alignment angle lies within 90°, the network trained by augmented DFA is roughly in the same direction as the BP would be. Here, we analyzed the alignment angle of the network with four hidden layers for the following conditions to scan $\eta$: $f(x)=\cos(x)$, $g(x)=\sin(x+\theta)$, where $\theta = 0, \pi/4, \pi/2, 3\pi/4, \pi$. Figure 2(d) shows the averaged alignment angle under the training as a function of $\eta$. For comparison, we also show the results for the case of BP by replacing $f'(x)$ with $g(x)$ in eq. (1). Their evolutions under the training are shown in Fig. 2(e) and (f). As can be seen, the alignment angles for the BP are significantly increased when the $\eta$ is apart from one. The alignment angle is beyond 90°, which reflects the test error increase shown in Fig. 2(a). On the other hand, the alignment angle for augmented DFA is highly robust to the $\eta$ value and smaller than 90°. These results suggest that we can train the deep physical network using inaccurate $f'(x)$ (or even using an alternative nonlinear function), which provides ease of physical implementations.

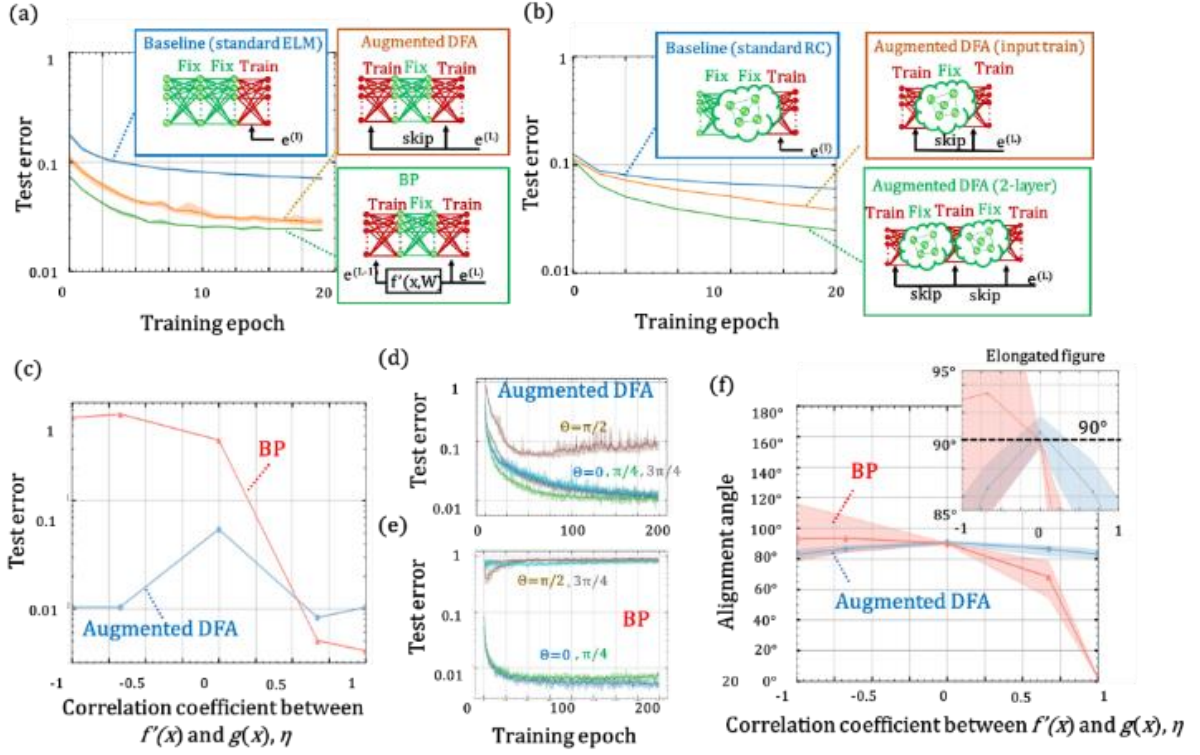

*Fig. 3.* Test accuracy for the (a) fully connected three-layer neural network with fixed frozen layer (multilayer ELM) and (b) for the one- and two-layer RC. As a baseline, the results for the standard ELM and RC approaches (i.e. read-out-only training) are also plotted in both figures. (c) Test error of four-layer RC trained as a function of $\eta$. Blue and red plots in (c) are the results for the model trained by BP and augmented DFA. $\eta$ was scanned by using $g(x)=\sin(x+\theta)$ with $\theta=0, \pi/4, \pi/2, 3\pi/4$, and $\pi$. (d) Averaged alignment angle under the training as a function of $\eta$. The colored region expresses the range within the minimum and maximum values. Evolution of alignment angles for layer 2 under training for (e) augmented DFA and (f) BP. Each experiment was repeated five times.

Let us discuss the effectiveness of the augmented DFA against the randomly fixed deep network [54] towards the application to physical RC and ELM. For this purpose, we investigated the performance for the deeply connected ELM and RC using MNIST task. Figure 3(a) shows the test accuracy as a function of training epoch for the deep neural network including the randomly fixed fully connected layers (deep ELM). In this experiment, we compared the performance for the three-layer network trained by using BP and DFA. The $f(x)$ and $g(x)$ are set to $\cos(x)$ and $\sin(x)$ to investigated the applicability of standard DFA algorithm at first. For the baseline, we also show only-readout training. As can be seen in the figure, we could train such a network by using DFA with almost the same accuracy as the one trained by BP, which were superior to one for readout-only training. Figure 3(b) shows the accuracy for test accuracy for the network including randomly fixed recurrent layers (deep RC). In this experiment, we compared the performance for the single-layer and two-layer RC layer with DFA training. The $f(x)$ and $g(x)$ are set to $\cos(x)$ and $\sin(x)$ to investigated the applicability of standard DFA algorithm. For the baseline, we also examined the single RC layer with readout-only training (the detailed image-processing scheme for deep RC is described in detail in Supplemental Material S3). As can be seen in the figure, we succeeded in training both the single-layer and stacked RC using DFA. The results in Fig. 3(a) and (b) suggest that DFA-based training is also effective in a deep neural network including passive layers. Importantly, although BP requires information about the passive layer, augmented DFA does not require it because we only need the random projection of the final error. Thus, it is highly suitable for black-box passive physical networks.

Next, we investigated robustness against the choice of $g(x)$ in augmented DFA algorithm using the deep RC with four hidden layers. Figure 3(c) and (d) show the test accuracy for the MNIST task and the average alignment angle as a function of $\eta$. We also plotted the results for the same network trained by BP for comparison by replacing $f'(x)$ to $g(x)$. The evolutions of alignment angle under the training are shown in Fig. 3(e) and (f). As can be seen, both the accuracy and alignment angle of the RC trained by augmented DFA are robust against the choice of $g(x)$, unlike the results for the BP training, which is the same trend in the results for the fully connected network. These results basically support the effectiveness of the augmented DFA approach even in RC. However, the alignment angle for the case with $\eta=0$ was larger than 90°; we should avoid this region to achieve better performance.

**Physical implementation**   Here, we show concrete examples of physical implementations for PNNs trained by augmented DFA, including diffractive ELM, nanophotonic NN, and optoelectronic deep RC. Among them, we demonstrate in this section a prototype hardware/software implementation of deep optoelectronic RC using an FPGA-assisted fiber-optic system. The rest of the demonstrations based on numerical simulations are described in Supplementary Material S1.

Up to now, various physical implementations of single-layer RC have been achieved by using a delayed dynamical system with a single nonlinear device [8, 17, 20-22]. By expanding this concept, we implement deep RC by cascading the delayed dynamical system. Figure 4(a) shows a schematic explanation for our constructed optoelectronic deep-RC benchtop. The equilibrium network topology is shown in Fig. 4(b). In this system, temporal input signals of the $l$th layer, $x_i^{(l)}(n)$, are masked and converted to $M_i^{(l)} x_i^{(l)}(n)$ using mask function $M_i^{(l)}$, where $i$ is the virtual node number ($i = 1, 2,…, N$, where $N$ is the number of virtual nodes in each reservoir layer). This operation generates quasi-random connections between inputs and virtual nodes. The masked inputs are converted to optical signals using an optical intensity modulator with time interval $\theta_{rc}$. Thus, the input signals are elongated to time interval $T = N\theta_{rc}$. The signals are introduced into a delay ring with a single nonlinear system, which acts as the reservoir layer. If we set the length of the delay ring $\tau$ as $\tau=T$, each node only couple with previous state of itself [means $\Omega$ in eq. (6) becomes diagonal matrix]. On the other hand, by choosing $\tau=(N+k)\theta_{rc}$, we can obtain a coupling between $x_i$ and $x_{i-k}$, which makes richer dynamics [55, 56]. Thus, we set the delay time as following desynchronized condition: $\tau=(N+1)\theta_{rc}$. The signals are directly detected by a single photodiode, and their discretized dynamic responses are considered as virtual nodes. These signals are converted to digital signals, which are stored in the memory. Then, they are considered as the next layer input signal $x_i^{(l+1)}(n)$. They are masked by $M_i^{(l+1)}$ and re-input to the RC system. The rest of the processing scheme is the same as in the previous layer. Since this scheme shares all the hardware components, the device architecture is simple, cost-effective, and easy to implement. Other possible photonic implementations are summarized in Supplemental Material S2.

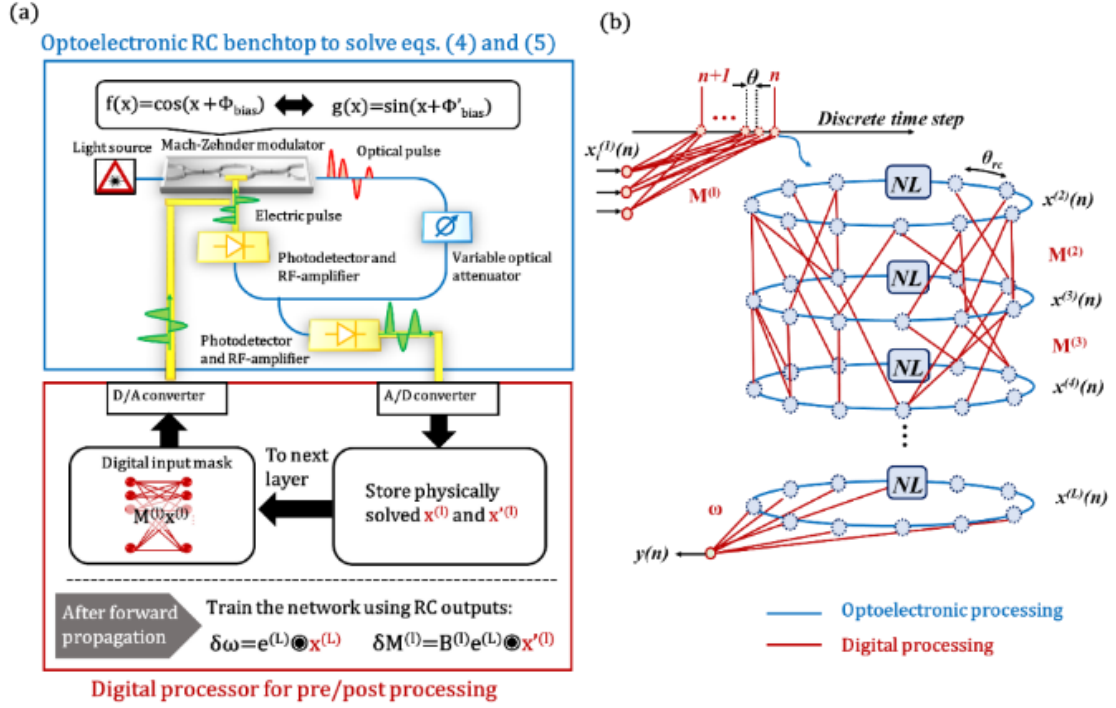

***Fig. 4.*** *(a) Schematic of the constructed optoelectronic deep RC. The input signals are masked by a digital processor and sent to the optoelectronic RC system to solve eqs. (4) and (5). The change in the nonlinearity from f(x) to g(x) is realized by applying bias to the Mach-Zehnder modulator. Based on physically solved x and x' values, the mask for each layer is updated. (b) Equilibrium network topology for constructed optoelectronic RC. Each reservoir layer shows ring topology since the RC system is composed of delay-based nonlinear fiber ring.*

As a nonlinear device, we employed a Mach-Zehnder interferometer, which provides the following activation: $f(x)=\cos(x+\Phi_{bias})$. Then, the obtained virtual node response $x_i^{(l+1)}(n)$ can be described as

$$x_i^{(l+1)}(n) = \cos\{\alpha x_{i-1}^{(l)}(n-1) + M_i^{(l)} x_i^{(l)}(n) + \Phi_{bias}\}, \qquad (4)$$

where $\alpha$ is feedback gain in the nonlinear delay ring. The operation in the next layer is the same as in the first layer. The outputs $y(n)$ are obtained from weighted summation of final layer output $x_i^{(L)}(n)$, as same as eq. (7) in Methods. Therefore, this stacked architecture of nonlinear delay-line-based oscillators can simply emulate the special type of deep RC. For the training, we need to calculate eqs. (8)-(10) in Methods. In particular, eq. (9) requires heavy computational costs because it includes recurrent computation with $O(N^2)$ operation with each time-step, the same as eq. (6) in the forward propagation case. However, in this system, eq. (9) can be solved by using the same optical system characterized by eq. (4). By setting the bias as ($\Phi'_{bias} + \pi/2$), we can generate an alternative nonlinear activation as follows:

$$x'^{(l+1)}_i(n) = sin\{\alpha x'^{(l)}_{i-1}(n-1) + M^{(l)}_i x'^{(l)}_i(n) + \Phi'_{bias}\}. \tag{5}$$

By scanning the $\Phi'_{bias}$ value, we can sweep the $\eta$ value from -1 to 1 to investigate the robustness against $\eta$. From digitally solved eqs. (8) and (10) and physically solved eq. (9), we can train the inherent parameters $M^{(l)}_i$ and ω. Note that we can solve eq. (8) using additional optical hardware such as optical random matrix operation [53], and this approach might accelerate the computational speed further.

Based on the above proposals, we constructed deep optoelectronic RC by combining a high-speed optoelectronic system developed for optical telecom with a highly programmable field programmable gate array (FPGA) with fast digital-to-analog/analog-to-digital converters (DAC/ADC). We also developed Pytorch/Python-compatible middleware for the ease of use of our device (see Supplementary Material S6). Using the benchtop, we examined the standard benchmark tasks. Figure 5(a) shows the layer dependence of the test accuracy of the MNIST task (the details of experimental setup are described in method section; the image processing scheme using deep RC is described in Supplemental Material S3). In this experiment, we set to the virtual node number $N=404$, $\alpha=0.9$, and $\Phi'_{bias}$ to 0 [i.e. $g(x)$ is equal to $f'(x)$ ideally]. $L=0$ means readout-only training; $L=1$ means both readin and readout training. As can be seen, the performance was improved by increasing the number of layers, as expected from the simulation. This result suggests the effectiveness of the augmented DFA algorithm for physical RC. Figure 5(b) shows the experimentally obtained test error as a function of correlation coefficient $\eta$. For comparison, the simulation results described in the previous section are also plotted. The robustness of test error to the change in $g(x)$ exceeded that expected from the simulation. This result might originate from the noise and setting error of the physical system.

The obtained test accuracy for MNIST, Fashion MNIST, and CIFAR-10 are summarized in Table I. The reported values for previous photonic DNN implementations and the RC based on other physical dynamics are also summarized in the Table. I [32], [33], [57]-[63]. In spite of the simplicity of our delay-based hardware implementation, we achieved competitive performance for all the examined tasks with the state-of-the-art large-scale benchtop. This supports the effectiveness of our approach.

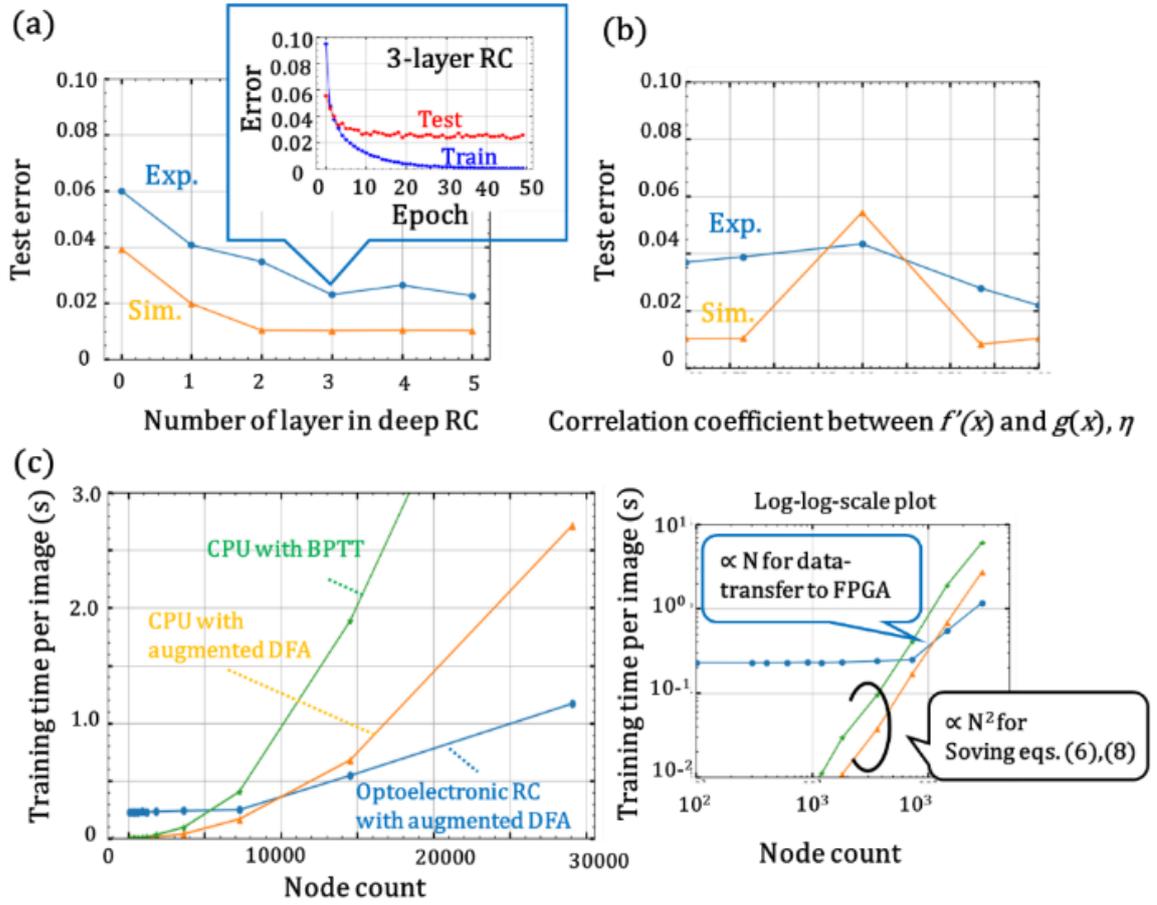

*Fig. 5. (a) Train accuracy as a function of layer number of RC. Blue and orange plots indicate the results for the constructed benchtop and simulation on a CPU. Inset of (a) shows train and test accuracy under the training. (b) Test error as a function of η. The accuracy is robust against shape of g(x). (c) Processing time as a function of node count. The same results with a log-log-scale plot are displayed on right side. The test was repeated three times.*

To evaluate the efficiency of our scheme, we measured the computational time for training our system. Although previous studies also compared the processing time of PNNs, they basically only evaluated the matrix operation time [e.g. $\mathbf{x}^{(l+1)} = f(\mathbf{W}^{(l)}\mathbf{x}^{(l)})$]. However, PNNs require many additional operations such as data transfer to the physical system, training based on simulation model, DAC/ADC operations, and pre- and post-processing for physical processing. Thus, the advantage for whole processing time was still under the consideration. Thanks to our constructed physically accelerable algorithm and its FPGA-assisted hardware implementation with full Pytorch/Python compatible software, we can evaluate of the whole processing time of our device, including the training time. As a first step, we investigated the processing time of PNNs by changing the node count, since the advantage of the physical RC approach lies in the acceleration to solve eqs. (6) and (8), requiring $O(N^2)$ computational costs.

|  |  | Data set | | | Training method |
| --- | --- | --- | --- | --- | --- |
|  |  | MNIST | Fashion MNIST | Cifar-10 |  |
| PNN based on photonics | **Deep RC [sim.] (This study)** | **99.03% (w/o preprocessing)** | **86.31%** | **56.85%** | **Augmented DFA (with optoelectronic hardware)** |
|  | **Deep RC [exp.] (This study)** | **97.80% (w/o preprocessing)** | **85.91%** | **47.83%** |  |
|  | Diffractive photonic DNN (ref [32], [58]) | 96.6 % [exp.] (w/o preprocessing) | 84.6% [exp.] | 44.4% [exp.] | BP (on external standard computer) |
|  | On-chip photonic DNN (ref [62]) | 95.3% [exp.] (w/o preprocessing) | Not reported | Not reported |  |
|  | Large-scale Photonic RC (ref [57]) | 97.15% [sim.] (w/o preprocessing) 98.90% [exp.] (with preprocessing) | Not reported | Not reported | Linear regression (on external standard computer) |
|  | On-chip Photonic RC (ref [33]) | 91.3% [exp.] (w/o preprocessing) | 70.1% [exp.] | Not reported |  |
| RC based on other physical dynamics | Spintronic RC (ref [61]) | 87.6% [sim.] (w/o preprocessing) | Not reported | Not reported | Linear regression (on external standard computer) |
|  | Memristor RC (ref [60]) | 88.1% [exp.] (w/o preprocessing) | Not reported | Not reported |  |
|  | Diffusive memristor RC (ref [63]) | 83% [exp.] (w/o preprocessing) | Not reported | Not reported |  |
|  | Self-organized nanowire RC (ref [59]) | 90.4% (w/o preprocessing) | Not reported | Not reported |  |

*Table I.* Scores for MNIST, Fashion MNIST and Cifar-10 for our device. The node counts are set to 600 and layer number were 5. For comparison, the reported scores for the PNNs based on the photonics and the RC based on other dynamics are also shown [32], [33], [57]-[63]. The features of our approach are high performance even in the simple optical implementation and training based on optoelectric dynamics, which can accelerate both inference and training speed. Sim. and exp. means simulation and experimental results. Pre-processing means image processing before the physical inputs such as Gabor filter, which can enhance the performance [57]. For the fair comparison, we focused on the results without pre-processing.

Figure 5(c) shows the measured training time per image for our constructed optoelectronic RC benchtop. For comparison, we also show the results for augmented DFA and BP training on the CPU (Intel Xeon W-2145, 8 cores, 3.7-GHz clock). The budget of processing time for the RC benchtop is broken down as follows: FPGA processing (data-transfer, memory allocation, and DAC/ADC) of ~92%, digital processing of ~8% for pre/post processing, including the time for solving of eqs. (7), (8), and (10), and optoelectronic processing time of ~0.02% for solving eqs. (6) and (8). Thus, the processing time is denoted as the data-transfer and allocation to the FPGA, whereas the calculation on the CPU does not require data-transfer, and its speed is denoted as the time to solve eqs. (6) and (8). As can be seen, the CPU shows $O(N^2)$ trends against the node count, the benchtop shows $O(N)$, which is due to the data-transfer bottleneck. (We need $O(N)$ memory on the FPGA board, but the memory size on the FPGA is limited. Thus, we need to increase the number of data-transfers by reducing minibatch size, which results in a linear time increment against $N$.) The physical acceleration beyond the CPU was observed at $N$ ~5,000 and ~10,000 for the BP and augmented DFA algorithm, respectively. This trend is same order with the previous estimation on the forward propagation of a photonic reservoir computer [52]. To the best of knowledge, this is the first comparison of the whole training process and the first demonstration of physical training acceleration using PNNs.

**Discussion**

**Augmentability to other physical systems** In this study, we have verified the effectiveness of our approach through physical experimentations using an optoelectronic delay-based implementation. The remaining question is an applicability to other systems. To answer it, we performed numerical simulations using a widely investigated photonic neural network and revealed the effectiveness of our approach even in complex-valued diffractive networks and nanophotonic unitary network networks (see Supplementary Material S1). In addition, our demonstrated delay-based RC is highly suitable and experimentally demonstrated for various physical systems. The major difference in other physical systems is the nonlinearity in eq. (4), which is sometimes difficult to identify accurately. However, as described above, our method is highly robust to $g(x)$, which suggest the algorithm is effective for such cases.

**Limitation and scalability of proposed algorithm** It was reported that the DFA algorithm was difficult to apply to convolutional neural networks (CNNs) [64]. Thus, the proposed method may be difficult to apply to convolutional PNNs [31], [58], [62] and CNN-based architectures such as AlexNet [65]. On the other hand, a recent study revealed that only a full-connection network can achieve state-of-the-art performance [66], which suggests the DFA-applicable architecture is comparable to the CNN-based one. In fact, it has also been reported that DFA can train modern network architectures without a convolution layer [51]. They suggest that our algorithm might work on much more complex tasks. Regarding the deeply connected passive physical system, its performance for complex tasks is still under investigation. So far, interestingly, it has been reported that the RC-based transformer model (transformer with a fixed layer trained by BP) works well [67]. As the transformer can be applied to many practical models, our deep RC scheme might scale to more advanced models. Further investigation will be performed in future. Regarding the physical system, the major issue for constructing a deep network is its intrinsic noise. Here, we investigated the effect of noise by numerical simulation (see Supplemental Material S7). We found the system is robust to the noise.

**How to select alternative nonlinearity**    In this work, we introduced alternative activation for the training. Although $g(x)$ is basically an arbitrary function, we should avoid the function near the $\eta=0$. One simple method to avoid it is to use $g(x)=\sin(x+\theta)$. By scanning $\theta$, we can sweep the $\eta$ value for various functions and find a good solution. In addition, this nonlinearity is suitable for some physical implementations and, as shown in this article, we can accelerate its operation. Another approach is to use optimization problems such as a genetic algorithm (GA). Although a GA is hard to implement in a physical system, we can find a good solution for complex physical nonlinearity. An example of optimization is shown in Supplemental Material S4.

**Further physical acceleration**    Our physical implementation confirmed the acceleration of recurrent processing for RC with a large-scale node count. However, its advantage is still limited, and further improvement is required. As mentioned in the Results section, the processing time of our current prototype is denoted as the data-transfer and memory allocation to the FPGA. Thus, the integration of the whole processing into the FPGA will improve the performance much more, with the sacrifice of experimental flexibility. In addition, in future, an on-board optics approach will reduce of transfer cost drastically. Large-scale optical integration and on-chip integration will further improve the optical computing performance itself.

**Conclusion**

In summary, we presented physical deep learning by augmented DFA. By augmenting the DFA formula with arbitrary nonlinearity representation, we revealed that the performance is robust to the choice of $g(x)$. Thus, we no longer need to simulate the physical system accurately for training. As the proposed method is based on parallel random projection with arbitral nonlinear activation, we can execute the computation for the training on a physical system in the same manner as physical ELM or the RC concept. We demonstrated proof-of-concept using a hierarchically connected optoelectronic recurrent neural network. Although our benchtop is simple and easy to apply to various physical platforms with only software-level updates, we achieved performance comparable with that of complex state-of-the-art large-scale systems. Moreover, we compared the whole processing time, including digital processing, and found the possibility of physical acceleration of the training procedure. Our approach provides practical alternative solutions for training and the acceleration of the large-scale physical neuromorphic computation.

**Methods**

**Augmented DFA in RC**  The forward propagation of RC is given by

$$\mathbf{x}^{(l)}(n) = f\{\mathbf{\Omega}^{(l)}\mathbf{x}^{(l)}(n-1) + \mathbf{M}^{(l)}\mathbf{x}^{(l-1)}(n)\}, \tag{6}$$

where $\mathbf{x}^{(l)}$ is the internal state of the $l$th reservoir layer, $\mathbf{M}$ is the connection between ($l$-1)th and $l$th reservoir layers (called a mask function), $\mathbf{\Omega}$ is the fixed random internal connection in the $l$th reservoir layer, and $n$ is the discrete time step. The final output $\mathbf{y}$ is obtained by

$$\mathbf{y}(n) = \boldsymbol{\omega}\mathbf{x}^{(L)}(n) \tag{7}$$

where $\boldsymbol{\omega}$ is the output weight. Based on the update rule for the DFA in a recurrent neural network, gradients $\delta\mathbf{M}^{(l)}$ and $\delta\boldsymbol{\omega}$ can be calculated by using the following equations [68].

$$\delta\mathbf{M}^{(l)} = \frac{\partial E}{\partial \mathbf{M}^{(l)}} = \frac{\partial \mathbf{x}^{(l)}(n)}{\partial \mathbf{M}^{(l)}}\frac{\partial E}{\partial \mathbf{x}^{(l)}(n)} = -\mathbf{x}^{(l),\mathrm{T}}(n)\big[\mathbf{B}^{(l),\mathrm{T}}\mathbf{e}^{(L)}(n)\big]\odot\mathbf{x}'^{(l)}(n), \tag{8}$$

$$\mathbf{x}'^{(l)}(n) = g(\mathbf{\Omega}^{(l)}\mathbf{x}'^{(l)}(n-1) + \mathbf{M}^{(l)}\mathbf{x}'^{(l-1)}(n)) \tag{9}$$

$$\delta\boldsymbol{\omega} = \frac{\partial E}{\partial \boldsymbol{\omega}} = -\mathbf{e}^{(L)}(n)\odot\mathbf{x}^{(L)}(n) \tag{10}$$

where $g(x)$ is the arbitrary function, and $\mathbf{e}^{(L)}$ is the error at the final layer (see Supplemental Material S5 for the derivation). In the standard RC framework, only $\boldsymbol{\omega}$ is trained by linear regression. On the other hand, our algorithm enables the training of both $\boldsymbol{\omega}$ and $\mathbf{M}^{(l)}$ for each layer. In a typical physical RC system, the operation of $\mathbf{M}^{(l)}\mathbf{x}(n)$ is executed by digital pre-processing. Therefore, the training of $\mathbf{M}$ is familiar to the physical implementation. Although the training of $\mathbf{M}^{(l)}$ can be executed by BP, it requires prior knowledge of $\mathbf{\Omega}^{(l)}$, $\mathbf{M}^{(l)}$, and $f$. This approach also requires the BPTT, which incurs serious computational costs. Meanwhile, augmented DFA does not require any knowledge about the physical system. By comparing with the augmented DFA for standard fully connected layers, we need to calculate eq. (9) additionally. However, this output can be calculated by using a physical system.

**Correlation coefficient**  To discuss the distance between $f'(x)$ and $g(x)$ quantitatively, we

measure the correlation coefficient $\eta$ between $f'(x)$ and $g(x)$, defined as

$$\eta = \frac{\int_{-e}^{e}\{f'(x)-\overline{f'(x)}\}\{g(x)-\overline{g(x)}\}dx}{\sqrt{\int_{-e}^{e}|f'(x)-\overline{f'(x)}|^2 dx}\sqrt{\int_{-e}^{e}|g(x)-\overline{g(x)}|^2 dx}}, \tag{11}$$

where $e$ is the natural logarithm, and the overlines mean the average. In order to discuss eq. (11) in the bounded range where the data is distributed, the integration range is set to [-$e$: $e$].

**Optoelectronic benchtop** In this device, datasets on the standard computer are transferred to the FPGA (Xilinx Zynq, Ultra-scale) via an ether cable. The matrix-operation of $\mathbf{M}^{(l)}\mathbf{x}^{(l)}$ is executed on the FPGA. Then, the signals are sent to the DAC (3-GHz, 4GSa/s, 8-bit resolution) on the FPGA. The analog electrical signals are converted to the optical intensity by using a LiNbO$_3$-based Mach-Zehnder modulator [Thorlabs LN05FC, 32-GHz bandwidth (BW)]. After transmitting the optical fiber-based delay line, the signal is detected by a photodetector (PD) [Finisar XPRV2022, 33-GHz BW]. The detected signals are amplified by a radio-frequency (RF) amplifier [SHF-S807C (SHF), 50-GHz BW:]. The internal dynamics are received by the PD and RF amplifiers via a 1:1 splitter. The optical signal is converted to an electrical signal by the PD, they sampled by the ADC on the FPGA. The received signals are reintroduced into the optoelectronic reservoir for the next layer calculation. After the forward propagation processing [eq. (4)] of each minibatch, we changed the bias condition from $\Phi_{bias}$ to $\Phi'_{bias}$ to change the nonlinearity from $f(\mathrm{x})$ to $g(\mathrm{x})$. Then, the same operation as the above-described forward propagation was re-executed to solve eq. (5). After the operation, augmented-DFA-based training was done on the CPU using the outputs from the optoelectronic processing. The optical system is configured to have a ring topology when the number of nodes is N=3636 and the sampling rate is S = 4GSa/s. The sampling rate can be changed under the constraint $S = S_{max}/k$, where $S_{max}$ = 4GSa /s, and $k$ is a natural number. The number of nodes can be changed by controlling $S$ under the condition $NS = constant$. The feedback gain $\alpha$ (spectral radius) can be controlled by changing the variable optical attenuator (VOA) value. All the above-described processing is implemented on the Pytorch-compatible software interface described in Supplemental Material S6. Thus, we can use this optoelectronic RC like a standard CPU or GPU (in Python code, the optoelectronic device can only be described as *device="oe_rc"*). The bottleneck of the computational speed is determined by the sampling rate of the DAC/ADC. The node increments up to 29,088 displayed in Fig. 5(c) is realized by the node-reuse scheme proposed by Takano *et al*. [69], which enable virtual node increments beyond the distance limitation of the delay ring.

**Acknowledgement**

The authors are grateful to Mr. F. Sugimoto for his support on the hardware experiment. The authors would like to thank Mr. T. Tamori for his support on the software implementation.

**Author contributions**

M. N, K. I, and K.N conceived the basic concept of the presented physical deep learning method. M. N and K. I performed the numerical simulations. M. N. constructed the optoelectronic benchtop and executed the optical experiment. M. N and K. T developed the FPGA-based electric interface and Pytorch-based software implementation for the experiment. T. H, Y. K, and K.N supervised the project. M.N. wrote the initial draft of the manuscript. All the authors discussed the results and contributed to write the manuscript.

**Corresponding author**

Correspondence to Mitsumasa Nakajima, Katsuma Inoue, and Kohei Nakajima.

**Competing interests**

The authors declare no competing financial interests.

**Figure caption**

**Fig. 1.** (a) Schematics of physical neural networks (PNNs). Training sequence of PNN with (b) backpropagation, and (c) augmented biologically plausible training called direct feedback alignment (DFA). Augmented DFA enables parallel, scalable, and physically accelerable training of deep physical networks based on random projection with alternative nonlinearity $g(x)$.

**Fig. 2**. (a) Test error distribution of four-layer fully connected neural network as a function of the correlation coefficients between $f'(x)$ and $g(x)$, $\eta$. Blue and red boxplots in (a) are the results for the model trained by BP and augmented DFA. $\eta$ was scanned by using various $g(x)$:$\sin(x)$, $\cos(x)$, triangle(x), tanh(x) functions and random Fourier series. The boxes in the figure indicate the first and third quartile of the data distribution. The whiskers mean the minimum and maximum values except the outlier. Training curve for (b) augmented DFA and (c) BP with random Fourier series. Color differences indicate differences in activation function [$f(x)$=$\cos(x)$, $\tanh(x)$] and differences in the random seed to generate $g(x)$. (d) Averaged alignment angle under the training as a function of $\eta$. The colored region expresses the range within the minimum and maximum values. Evolution of alignment angle for layer 2 under training for (e) augmented DFA and (f) BP. Each experiment was repeated five times.

**Fig. 3.** Test accuracy for the (a) fully connected three-layer neural network with fixed frozen layer (multilayer ELM) and (b) for the one- and two-layer RC. As a baseline, the results for the standard ELM and RC approaches (i.e. read-out-only training) are also plotted in both figures. (c) Test error of four-layer RC trained as a function of $\eta$. Blue and red plots in (c) are the results for the model trained by BP and augmented DFA. $\eta$ was scanned by using $g(x)=\sin(x+\theta)$ with $\theta=0, \pi/4, \pi/2, 3\pi/4$, and $\pi$. (d) Averaged alignment angle under the training as a function of $\eta$. The colored region expresses the range within the minimum and maximum values. Evolution of alignment angles for layer 2 under training for (e) augmented DFA and (f) BP. Each experiment was repeated five times.

**Fig. 4.** (a) Schematic of the constructed optoelectronic deep RC. The input signals are masked by a digital processor and sent to the optoelectronic RC system to solve eqs. (4) and (5). The change in the nonlinearity from $f(x)$ to $g(x)$ is realized by applying bias to the Mach-Zehnder modulator. Based on physically solved $x$ and $x'$ values, the mask for each layer is updated. (b) Equilibrium network topology for constructed optoelectronic RC. Each reservoir layer shows ring topology since the RC system is composed of delay-based nonlinear fiber ring.

**Fig. 5.** (a) Train accuracy as a function of layer number of RC. Blue and orange plots indicate the results for the constructed benchtop and simulation on a CPU. Inset of (a) shows train and test accuracy under the training. (b) Test error as a function of $\eta$. The accuracy is robust against shape of $g(x)$. (c) Processing time as a function of node count. The same results with a log-log-scale plot are displayed on right side. The test was repeated three times.

**Table I.** Scores for MNIST, Fashion MNIST and Cifar-10 for our device. The node counts are set to 600 and layer number were 5. For comparison, the reported scores for the PNNs based on the photonics and the RC based on other dynamics are also shown [32], [33], [57]-[63]. The features of our approach are high performance even in the simple optical implementation and training based on optoelectric dynamics, which can accelerate both inference and training speed. Sim. and exp. means simulation and experimental results. Pre-processing means image processing before the physical inputs such as Gabor filter, which can enhance the performance [57]. For the fair comparison, we focused on the results without pre-processing.

# Supplemental Material for Physical Deep Learning with Biologically Plausible Training Method

Mitsumasa Nakajima[1], Katsuma Inoue[2], Yasuo Kuniyoshi[2,3], Kenji Tanaka[1], Toshikazu Hashimoto[1], and Kohei Nakajima[2,3]

1. NTT Device Technology Labs., 3-1 Morinosato-Wakamiya, Atsugi, Kanagwa, Japan

2. Graduate School of Information Science and Technology, The University of Tokyo

3. Next Generation Artificial Intelligence Research Center, The University of Tokyo

## S1. Application to another photonic system

**Unitary nanophotonic deep neural network** As an example of augmented DFA for a physical neural network, we show a numerical simulation of a unitary neural network composed of a nanophotonic unitary processor [1,2] as shown in Fig. S1(a). The nanophotonic processor is composed of arrays of $2 \times 2$ unitary operator $\mathbf{R_{ij}}$ based on a Mach-Zehnder interferometer (MZI). Any $N \times N$ unitary matrix $\mathbf{U}$ can be decomposed as a product of $\mathbf{R_{ij}}$ and diagonal matrix $\mathbf{D}$, such that $\mathbf{U} = \mathbf{D} \prod_{i=2}^{N} \prod_{j=1}^{i-1} \mathbf{R_{ij}}$, where $\mathbf{R_{ij}}$ is defined as the $N$-dimensional identity matrix with the elements $R_{ii}$, $R_{ij}$, $R_{ji}$, and $R_{jj}$ replaced as follows:

$$\begin{pmatrix} R_{ii} & R_{ij} \\ R_{ji} & R_{jj} \end{pmatrix} = \begin{pmatrix} e^{i\varphi}\cos\theta & -e^{i\varphi}\sin\theta \\ \sin\theta & \cos\theta \end{pmatrix}, \tag{S1}$$

where $\varphi$ and $\theta$ are the phases in the $2 \times 2$ unitary operator, which correspond to the phases in the MZI. Therefore, the MZI array can compose arbitral unitary operation. By adding nonlinear activation $f(x)$, the unitary neural network, described as $\mathbf{x}^{(l+1)} = f(\mathbf{U}^{(l)}\mathbf{x}^{(l)})$, can be composed by using the optical system shown in Fig. S1(a), where $\mathbf{U}^{(l)}$ is the unitary weight matrix and $\mathbf{x}^{(l)} \in \mathbb{C}$ is the complex-valued input/output electric fields. This arrangement is energy-efficient because the unitary operation does not require gain and loss of input light. The gradient of unitary matrix $\delta \mathbf{U}^{(l)}$ can be described by

$$\delta \mathbf{U}^{(l)} = \frac{\partial E}{\partial \mathbf{U}^{(l)}} = -\mathbf{x}^{(l),\mathbf{T}}[\mathbf{B}^{(l),\mathbf{T}}\mathbf{e}^{(L)}] \odot g(\mathbf{x}^{(l)}), \tag{S2}$$

As the considered device is composed of the unitary matrix operator and trainable parameters, random projection matrix **B** is sampled from complex-value. From eq. (S.2), the gradient of phases in the MZI ($\delta\boldsymbol{\theta}^{(l)}$ and $\delta\boldsymbol{\varphi}^{(l)}$) can be obtained from the relationships $\delta\boldsymbol{\theta}^{(l)}=(\partial \mathbf{U}^{(l)}/\partial\boldsymbol{\theta}^{(l)})\delta\mathbf{U}^{(l)}$ and $\delta\boldsymbol{\varphi}^{(l)}=(\partial \mathbf{U}^{(l)}/\partial\boldsymbol{\varphi}^{(l)})\delta\mathbf{U}^{(l)}$, where gradient $(\partial \mathbf{U}^{(l)}/\partial\boldsymbol{\varphi}^{(l)})$ and $(\partial \mathbf{U}^{(l)}/\partial\boldsymbol{\theta}^{(l)})$ can be solved by the adjoint sensitivity method. Therefore, we can obtain $\delta\theta$ and $\delta\varphi$ using the random projection of $\mathbf{e}^{(L)}$, the phase gradient against **U** ($\partial \mathbf{U}/\partial\boldsymbol{\theta}$ and $\partial \mathbf{U}/\partial\boldsymbol{\varphi}$), and arbitrary nonlinear projection $g(x)$. In addition, by introducing random unitary weight $\mathbf{B}_{unitary}^{(l)}$ as a random projection matrix in eq. (S.2), we can compute $\delta\mathbf{U}^{(l)}$ by using an optical system with almost the same configuration as that for forward propagations [see Fig. S1(b)]. In this system, we can execute unitary random projection of $\mathbf{e}^{(L)}$ using a photonic unitary processor by setting the random MZI phases. The Hadamard products with $g(x)$ are executed by using a beam splitter and optoelectric modulators based on an MZI whose nonlinear activation is described as $g(x) = \sin(x+\Phi_{bias})$, where $\Phi_{bias}$ is the control bias of the MZI. As described in the main text, we should avoid the point of $\eta = 0$ [i. e. $g(x)$ is near the inverse function of $f'(x)$]. By scanning $\Phi_{bias}$, we can avoid this condition. The computational cost of eq. (S.2) is O ($N^2L$). Therefore, there are advantages of optical implementation in computation speed and energy-consumption, the same as in forward propagation.

As a demonstration, we executed the numerical simulation for the MNIST task considering the optical setup shown in Fig. S1(a) and (b). We assume an $L$-layered unitary neural network with 64 nodes, where $L$ is scanned from 1 to 10. Thus, each network is composed of a 64 × 64 MZI array. For the nonlinear function, we selected the optoelectric nonlinearity $f(x)=\tanh(|x^{(l)}|^2)$ assuming the intensity detection and electrical tanh activation [3]. Since differential of tanh(x) is even function [$\{\tanh(x)\}'=1/\cosh^2(x)$], we use odd function of $g(x)=\cos(x)$ to avoid the point near $\eta = 0$. This condition can be realized by setting $\Phi_{bias}=\pi/2$. As the MZI has only 64 input ports, the input image was reshaped into an 8 × 8 size, and they were arranged to a 64-sized one-directional vector. The layer dependency of test error for the MNIST task is shown in Fig.S1(c). For comparison, we also plot the results using the BP training with $f'(x)$ replaced with $g(x)$. As can be seen in the figure, we could successfully train the nanophotonic unitary neural network even using augmented DFA. The results also suggest that the DFA approach is effective for complex-valued unitary neural networks. The accuracy for the back-propagation training became worse when the network became deeper because $f'(x)$ is replaced to $g(x)$, which suggests that augmented DFA is more effective for deep networks. Although the achievable accuracies for relatively shallow networks were inferior to those for BP, it is still considered useful because we do not need to solve back-propagation and eq. (S.2) can be solved with an optical processor.

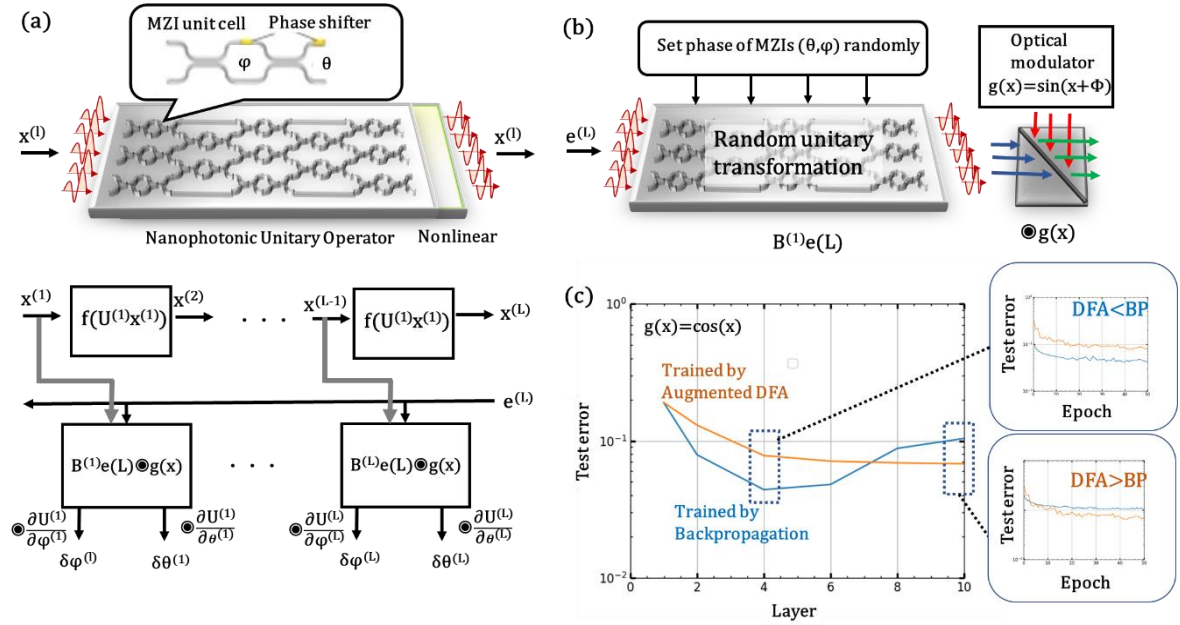

*Fig. S1. (a) Schematic illustration of deep unitary neural network composed of unitary nanophotonic operator. (b) Implementation of optical solver for random projection. (c) Numerical experimental test error as a function of the number of the network layer.*

**Diffractive deep neural networks**   As another demonstration of DFA in a photonic system, we show the inverse optimization of a diffractive deep neural network (D²NN). Figure S.2(a) shows the basic operation of the D²NN [4–6]. The D²NN is composed of diffractive optical elements (DOEs), which can modulate the optical wavefront by tuning local refractive index distributions. In this framework, we consider the complex-valued optical amplitude on the DOE plane as complex-valued neuron responses. The optical beam propagation within the D²NN can be described by using the same formula with fully connected complex-valued neural networks. The neuron weight corresponds to the complex transmission coefficient $T^{(l)}(x,y)$ of the DOEs on the $l$th DOE plane, which can be described as $T^{(l)}(x,y) = A^{(l)}(x,y)\exp\{j\Phi_i^{(l)}(x,y)\}$, where $A^{(l)}(x,y)$ and $\Phi^{(l)}(x,y)$ are the modulated amplitude and phase on the $l$th DOE plane. For a phase-only D²NN architecture, the amplitude $A^{(l)}(x,y)$ is assumed to be a constant value equal to 1. The D²NN can be applied to not only the machine learning accelerator demonstrated in main article and above section, but also to the inverse design of functional optics such as optical splitters and optical mode converters. Here, we numerically demonstrate the inverse design of 1:16 optical splitter using the DFA algorithm.

 In the demonstration, we assumed four-layer 3×3 mm² phase-only DOEs. The distance between DOEs were set to 50 mm. The input beam was assumed to be a single gaussian beam with the width of 100 μm. The target beam was set to 16 gaussian beams with the width of 100 μm [see Fig. S2(g)]. The grid size on the DOE plane was set to 5 μm, and the input wavelength was 1.5 μm. We assumed linear optics in this demonstration (there is no nonlinear activation). The forward propagation was calculated by the beam propagation method. For the optimization, we used the DFA algorithm. Figure S.2(b)-(e) show the designed phase pattern after the training. As can be seen, the designed pattern for each layer differ from each other, suggesting that each layer has a different optical function. Figure S.2(f) shows the mean square error (MSE) as a function of the training epoch. For comparison, the results with the adjoint method (continuous representation of BP) and final-layer-only training are plotted in the same figure. As can be seen, the MSE of both BP and DFA training was lower than that for the final-layer-only training. This suggests successful training in the former layer. Although the MSE for the model trained by DFA is slightly inferior to that for the BP, the optimized beam pattern well agrees with the target pattern. As described in the main article, DFA incurs less computational cost and is highly robust to the estimation error in a physical system. Thus, it may be useful for the fast and robust inverse design of physical systems, including $D^2NNs$. Further investigation remains as future work.

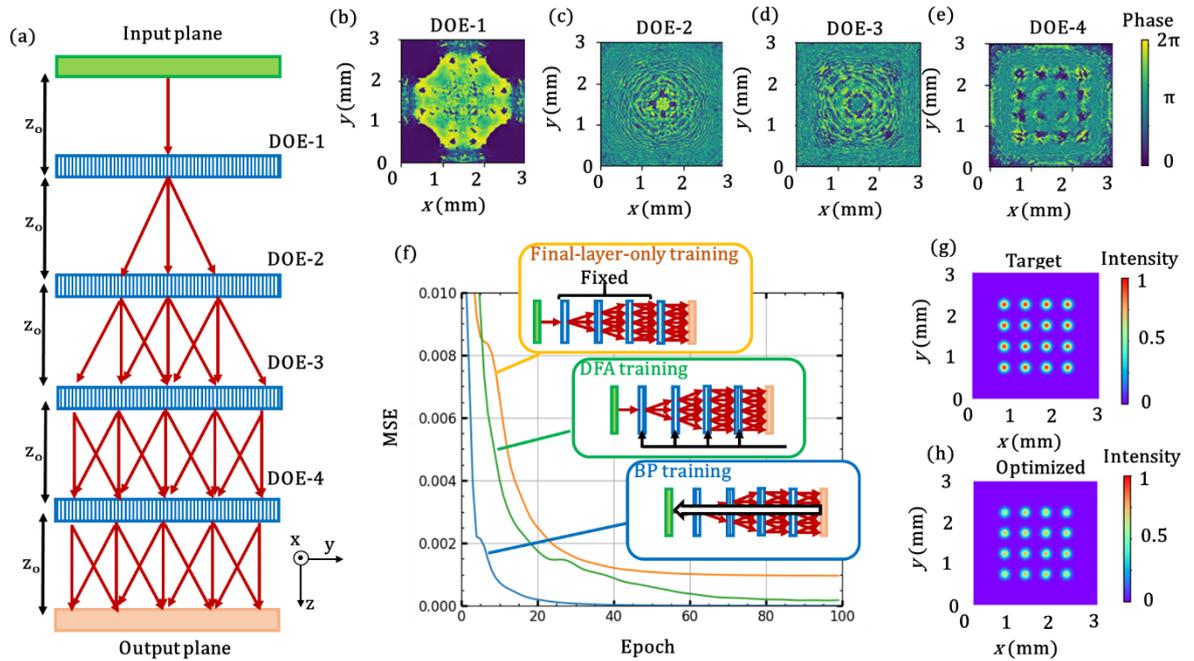

*Fig. S.2. (a) Schematic illustration of deep diffractive neural network composed of diffractive optical elements (DOEs). Optimized phase pattern for (b) first, (c) second, (d) third, and (e) fourth layer. (f) Numerical experimental mean square error (MSE) as a function of the number of network layers. (g) Target and (h) optimized optical intensity on the output plane.*

**S2. Possible implementation of deep RC based on delay-based dynamics**

Possible architectures for deep RC using delay-line-based implementations are listed in Fig. S3. The simplest approach is spatially stacked deep RC, which is illustrated in Fig. S3(a). In this architecture, the multiple photonic RC layers are cascaded layer by layer. This architecture can support real-time processing, and the reservoir parameters (e. g. feedback gain, mask function, output weight) can be tuned independently. However, physical architectures become complex because they require many optical components. Another approach is wavelength-division deep RC [Fig. S3(b)]. In this scheme, each wavelength corresponds to each layer for deep RC. Since the photonic circuits can be shared in this architecture, their complexity is reduced compared to spatially stacked deep RC. However, we still need multiple transmitters and receivers to implement multiple layers. The other approach is the time-sharing deep RC shown in Fig. S3(c). In this scheme, the optics is exactly the same as that in the conventional single-layer photonic RC. The memorized output signals are re-input to the RC system after the first epoch. Since all the RC layers share a single hardware component, the device architecture is simple and easy to construct.

|  | (a) Spatially stacked deep RC | (b) Wavelength multiplexed deep RC | (c) Time shared deep RC |
|---|---|---|---|
|  | 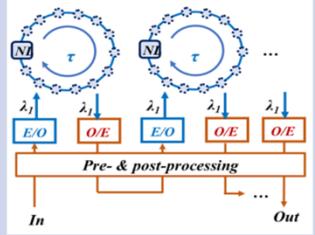 | 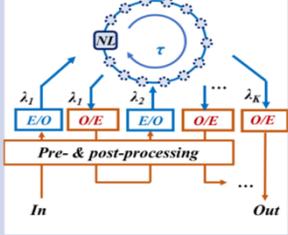 | 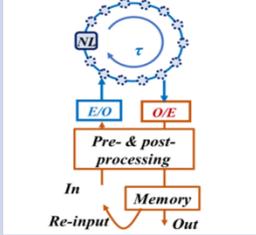 |
| Device complexity | Bad (Need many optical components) | Acceptable (Can share optical delay line) | Good (Can share every components) |
| Scalability | Acceptable (Limited by device complexity) | Good | Acceptable (Limited by device complexity) |
| Parameter tunability | Good (Can tune all parameters independently) | Acceptable (Need to share delay line characteristic) | Acceptable (Need to share delay line characteristic) |

*Fig. S3. Possible photonic implementations of deep RC and their comparison.*

### S3. Image recognition scheme

We describe the processing procedure for an image recognition task using deep reservoir computer. Figure S4 shows the processing method for V×H sized images using delay-based deep RC. The 1×V sized temporal input signals $\mathbf{x}^{(1)}(n)$ are masked by a V×N sized random input mask ($N$ is the number of reservoir nodes), and they are input to delay-ring-based reservoir layer. The discretized dynamic responses in reservoir layer are considered as virtual node of reservoir. The measured response is considered to be the next temporal inputs for the second layer, $\mathbf{x}^{(2)}(n)$. To classify the input digit image to 10-class (0, 1, 2,..,9), 10×1 dimensional outputs $\mathbf{y}(n')$ are obtained from weighted summation of virtual nodes in the final reservoir response $\mathbf{x}^{(L)}(n)$ as described in the following form:

$$y_k(n') = \sum_{j=0}^{H-1} \sum_{i=1}^{N} \omega_{ijk} x_i^{(L)}(n-j) \tag{S3}$$

where *n'* is time step for the output series with the time step interval of *H*, and $y_k$ is a *k*th element of output **y**.

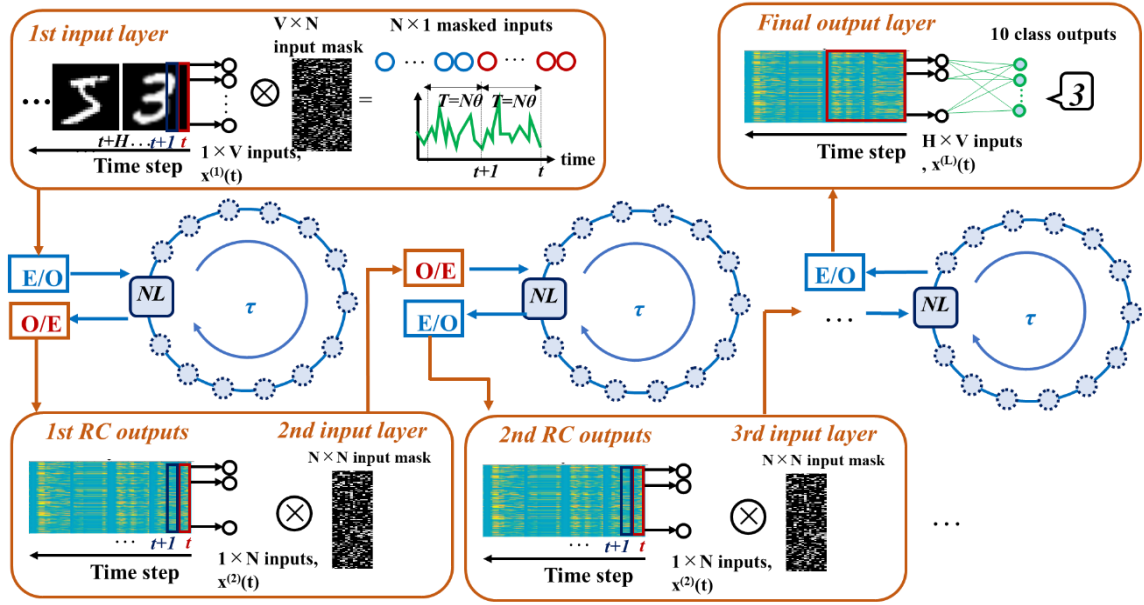

*Fig. S4. Image pprocessing method for delay-based deep reservoir computer.*

**S.4 Optimization of alternative nonlinearity using genetic algorithm**

In this work, we introduced alternative activation *g(x)* for the training. One question is how to find the optimal *g(x)* when we do not know the shape of *f(x)*. Here, we describe an approach utilizing a genetic algorithm (GA) called the particle swarm optimization (PSO) method [7]. Although it is hard to implement in a physical system, we can find a good solution for complex physical nonlinearity.

As a demonstration, we numerically examined four-layer fully connected networks with two nonlinear activation function. One is hyperbolic tangent (tanh). The other follows random Fourier series $f(x) = c_1 + \sum_{k=1}^{N} a_n \sin(kx) + b_n \cos(kx)$, where $a_k$, $b_k$, and $c_1$ are the random uniform coefficients sampled from $\mathbb{R} \in [-1:1]$ with *N* set to 4, and they are normalized by the relationship $|c_1| + \sum_{n=1}^{N} |a_n| + |b_n| = 1$. The node count in each hidden layer was set to 800. For the optimization, we initialized the g(x) using the Fourier series $g(x) = c_1 + \sum_{k=1}^{N} a_n \sin(kx) + b_n \cos(kx)$, where $a_k$, $b_k$, and $c_1$ are constrained the same as in the the case for *f(x)*. Based on the experiment, we updated the shape of *g(x)* utilizing the standard manner for the PSO method. For the optimization, the number of generations and groups were set to 20 and 128. Figure S5(a) shows the test error after the 20-epoch training as a function of the generation in the PSO algorithm. The shapes of $g(x)$ for the first generation and final (20th) generation are plotted in Fig. S5(b)-(e). As can be seen in Fig. S4(a), both the average and best error were decreased with the generation number, which suggests the success of our algorithm. In addition, the shape of $g(x)$ converged to near the shape of $f'(x)$. Interestingly, the achieved best error of 1.14% [for $f(x)$: tanh$(x)$] and 1.89% [for $f(x)$: random Fourier series] were almost comparable to the $g(x)=f'(x)$ case, though the converged $g(x)$ shape did not agree with the shape of $f'(x)$. This suggests the robustness of the proposed augmented DFA scheme.

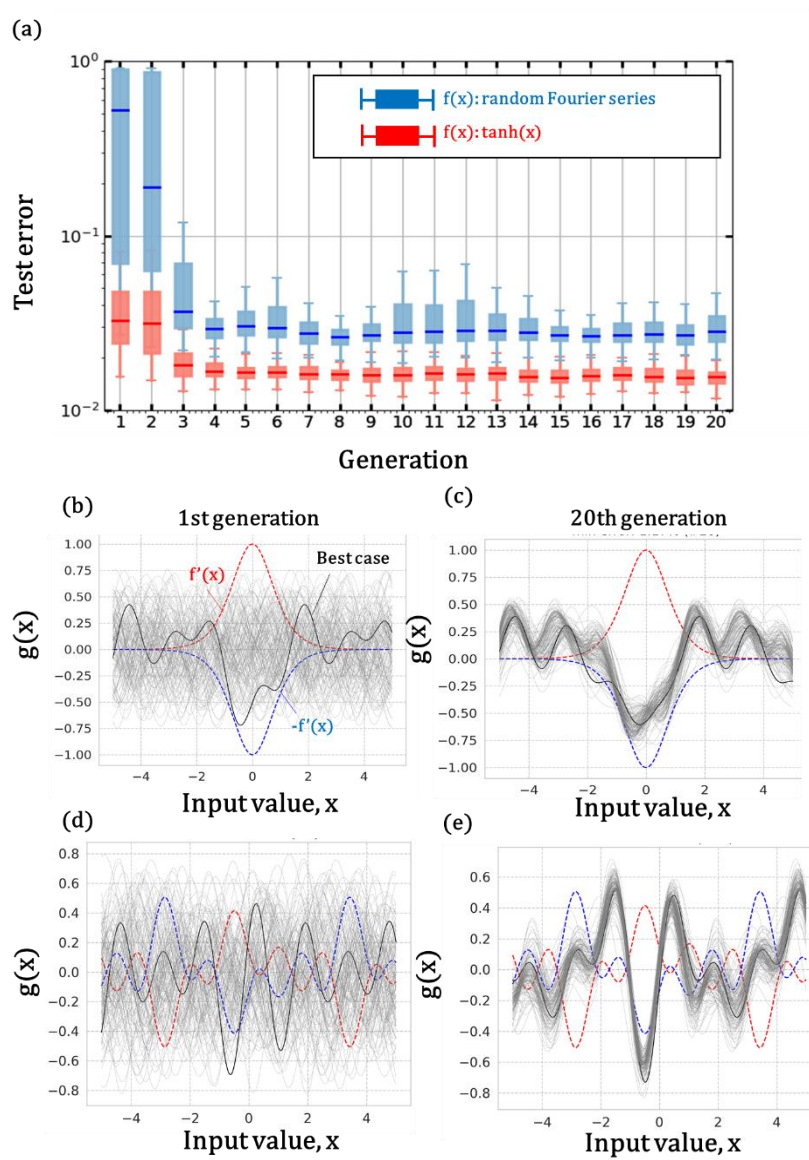

Fig. S5. (a) Test error as a function of generation in the PSO optimization. (b), (d) Initial g(x) shape for the case of (b) tanh and (d) random Fourier activation function. (c), (e) g(x) shape of 20th generation for the case of (b) tanh and (d) random Fourier activation function.

**S5 Augmented DFA in reservoir computer**

We describe the augmented DFA for deep RC. Based on the standard back-propagation (BP) algorithm, the gradient $\delta \mathbf{x}^{(l)}$ in eq. (6) can be described as

$$\frac{\partial E}{\partial \mathbf{x}^{(l)}(n)} = \{\Omega^{(l)} \frac{\partial \mathbf{x}^{(l)}(n+1)}{\partial \mathbf{x}^{(l)}(n)} + \mathbf{M}^{(l+1)} \mathbf{e}^{(l+1)}(n)\} \odot f'\{\mathbf{s}(n)\}, \tag{S4}$$

where

$$\mathbf{s}^{(l)}(n) = \Omega^{(l)} \mathbf{x}'^{(l)}(n-1) + \mathbf{M}^{(l)} \mathbf{x}'^{(l-1)}(n). \tag{S5}$$

By neglecting the first term in eq. (S4) and replacing the second term with a linear random projection of the error signal at the final layer, $\mathbf{e}^{(L)}(n)$, we can derive the following update rule:

$$\delta \mathbf{M}^{(l)} = \frac{\partial E}{\partial \mathbf{M}^{(l)}} = \frac{\partial \mathbf{x}^{(l)}(n)}{\partial \mathbf{M}^{(l)}} \frac{\partial E}{\partial \mathbf{x}^{(l)}(n)} = -\mathbf{x}^{(l),T}(n)[\mathbf{B}^{(l),T}\mathbf{e}^{(L)}(n)] \odot f'\{\mathbf{s}^{(l)}(n)\}. \tag{S6}$$

The same discussion for a recurrent network (RNN) with DFA training and its effectiveness have already demonstrated by Murray *et al* [8]. By augmenting the DFA algorithm to arbitrary nonlinearity $g(x)$, the same as the case for fully connected layers described in the main article, we finally obtain

$$\delta \mathbf{M}^{(l)} = -\mathbf{x}^{(l),T}(n)[\mathbf{B}^{(l),T}\mathbf{e}^{(L)}(n)] \odot g\{\mathbf{s}^{(l)}(n)\}. \tag{S7}$$

This equation is the same as eq. (8) in the main text.

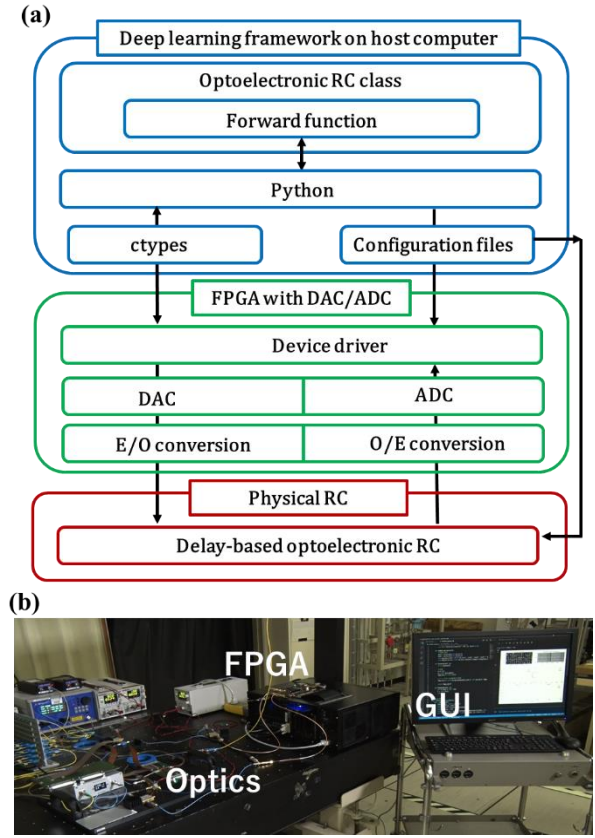

*Fig. S6. (a) Schematic processing flow of constructed optoelectronic reservoir computing system, (b) Image of constructed optoelectronic benchtop.*

## S6. Software and hardware interface for physical implementation

In order to execute the computation using optoelectronic-RC-based physical neural networks (PNNs), we need a functional connection between the analog-optical module and deep learning framework on a standard digital computer. Here, we developed the software/hardware interface shown in Fig.S6 (a). The image of constructed optoelectronic hardware for deep reservoir computing is also shown in Fig. 6(b). In this interface, we implemented a new class for the optoelectronic RC on Pytorch, which can be executed like a conventional network model, such as a fully connected layer and convolution layer. On the user-interface, we can name optoelectronic RC like a standard CPU or GPU by only describing *device="oe_rc"* in the python code. When the optoelectronic RC class is called, the physical parameters (e.g. feedback gain and node count) are tuned to realize a programmed setup through the configuration files. For the forward propagation, the data is passed to the FPGA via ctypes. To reduce the latency for the data transfer, we also employed an encoder and decoder whose operation speed is INT10 at 32.8 Gops/s and 131.0 Gops/s. After that, the data were aligned in a static random access memory (SRAM) and input to/output from the optoelectronic RC through a multichannel A/D converter and O/E converter. The obtained results correspond to the computation results of eqs. (4)

and (5) in the main article. These computational results are stored in the memory of the host computer. For the training phase, we update the weights using the stored results and the eqs. (8) and (9). These computations are also executed on the host computer.

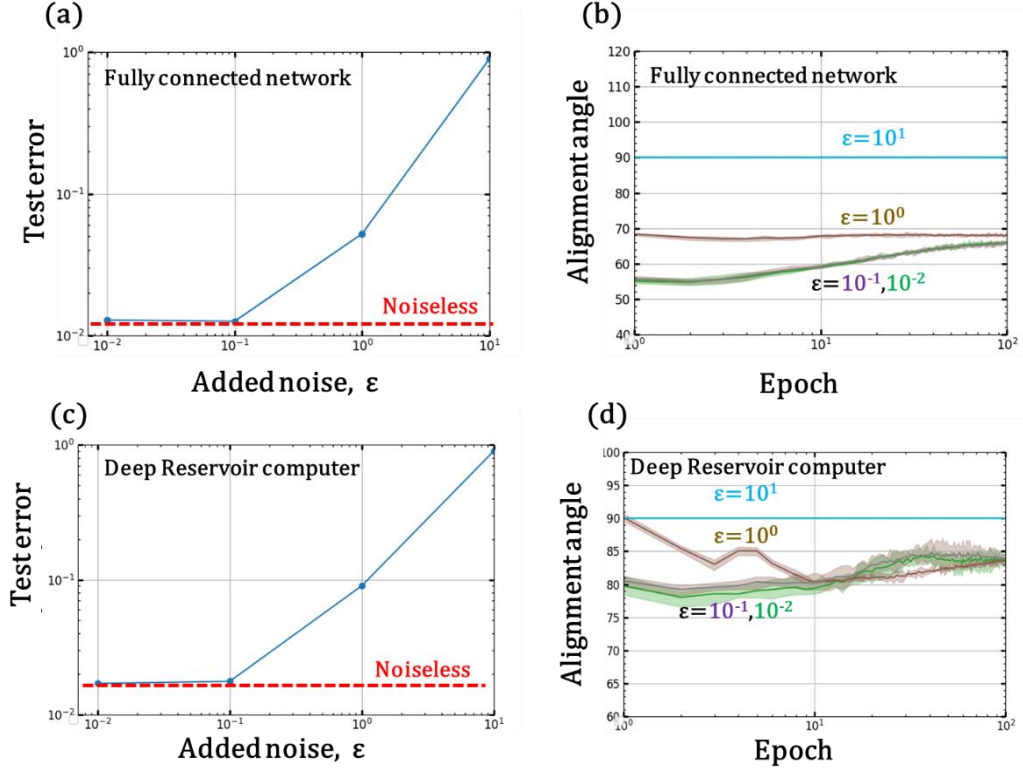

*Fig. S7. (a),(c) Test error for (a) four-layer fully connected network and (c) four¥layer deep reservoir computer as a function of added noise. (b),(d) Alignment angle for (b) four-layer fully connected network and (d) four-layer deep reservoir computer as a function of training epoch.*

**S7. Noise dependence**

For the physical implementation, the effect of noise is important because the analog physical signals (e.g., optical intensity, electric field, distortion of soft-body) are always affected by noises in the physical device (e.g., shot noise, vibration). Here, we investigated the effect of noise by considering the following noise addition. Here, we investigated the effect of noise for the DFA training using following equation.

$$f'(x) = f(x) + \varepsilon\sigma, \qquad (S8)$$

$$g'(x) = g(x) + \varepsilon\sigma, \qquad (S9)$$

where $\sigma$ is white gaussian noise, and $\varepsilon$ is noise intensity. By replacing the $f(x)$ and $g(x)$ values in eqs. (3), (6) and (9) in the main article with $f'(x)$ and $g'(x)$ described in eqs. (S8) and (S9), we numerically examined the effect of noise on the performance using the MNIST dataset. For this experiment, we employed a four-layer fully connected network and four-layer RC. $f(x)$ and $g(x)$ were set to $\cos(x)$ and $\sin(x)$, respectively. The other experimental conditions were the same as the ones for numerical experiment in the main article. Figure S7 (a) and (c) shows the test error as a function as a function of $\varepsilon$ for the fully connected network and deep RC. The $\varepsilon$ dependency of the alignment angle is plotted in the Fig. S7 (b) and (d). As can be seen, the performance and alignment angle below $\varepsilon=10^{-1}$ are almost same as in noiseless case. Our optoelectronic RC benchtop showed the signal-to-noise-ratio (SNR) of 24 dB, which corresponds to $\varepsilon=4 \times 10^{-2}$. Thus, we think that the effect of gaussian noise (e.g., noise from the amplifier and photodetectors) on the physical computation is relatively small.